\documentclass[a4paper,10pt]{article}

    \usepackage[latin1]{inputenc}
    \usepackage{graphicx}
    \usepackage{fancyheadings}

    \usepackage{amsmath}
    \usepackage{amsfonts}
    \usepackage{multirow}

\providecommand{\firstaffiliationmark}{ $^{\ast}$}

\providecommand{\micauthor}[3]{\index{\textsc{#2, #1}}#1 #2#3 \and }
\providecommand{\micinstitution}[4]{\small #1 #2 \\ #3 \\ Email: \texttt{#4} \\ \hfill \\ }
\providecommand{\institutions}[1]{\date{#1}}

\providecommand{\micfilename}[1]{#1}

\begin{document}

%%%%%%%%%%%%%%%%%%%%%%%%%%%%%%%%%%%%%%%%%%%%%%%%%%%%%%%%%%%%%%%%%%%%%%%%%%%
%
%      Your extended abstract starts here!
%
%%%%%%%%%%%%%%%%%%%%%%%%%%%%%%%%%%%%%%%%%%%%%%%%%%%%%%%%%%%%%%%%%%%%%%%%%%%

%Title of your extended abstract, e.g.

% Converted from Word by Pedro Borges, April 2001

\title{Genetic Algorithms for multiple objective vehicle routing}

% Your name(s): first name or initials, last name, affiliation mark, e.g.
\author{\micauthor{M.J.}{Geiger}{\firstaffiliationmark}
}

% Your affiliation(s), address(es), and e-mail(s) as they should appear
% in the proceedings volume, preceded by the affiliation mark used for the
% respective author, e.g.
\institutions{
%first institution
\micinstitution{\firstaffiliationmark}
    {Production and Operations Management}
    {Institute 510 - Business Administration\\
      University of Hohenheim}
    {mail@martingeiger.de}
}

\maketitle

% Your first section

\begin{abstract}
  The talk describes a general approach of a genetic algorithm for
  multiple objective optimization problems. A particular dominance
  relation between the individuals of the population is used to define
  a fitness operator, enabling the genetic algorithm to adress even
  problems with efficient, but convex-dominated alternatives. The
  algorithm is implemented in a multilingual computer program, solving
  vehicle routing problems with time windows under multiple
  objectives. The graphical user interface of the program shows the
  progress of the genetic algorithm and the main parameters of the
  approach can be easily modified. In addition to that, the program
  provides powerful decision support to the decision maker. The
  software has proved it´s excellence at the finals of the European
  Academic Software Award EASA, held at the Keble college/ University
  of Oxford/ Great Britain.
\end{abstract}

\section{The Genetic Algorithm for multiple objective optimization problems}

Based on a single objective genetic algorithm, different extensions
for multiple objective optimization problems are proposed in
literature
\cite{BackSchwefel1996,FonsecaFleming1993,Horn1997,TamakiKitaKobayashi1996}
All of them tackle the multiple objective elements by modifying the
evaluation and selection operator of the genetic algorithm. Compared
to a single objective problem, more than one evaluation functions are
considered and the fitness of the individuals cannot be directly
calculated from the (one) objective value.

\begin{figure}%[h]
\begin{center}
\includegraphics[width=5cm]{\micfilename{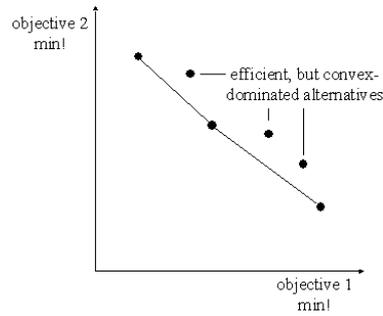}}   % Please use command
\vspace{-.5em} \caption{Efficient, convex-dominated alternatives}  % \micfilename
\label{pic1}
\end{center}
\end{figure}

Efficient but convex-dominated alternatives are difficult to obtain by
integrating the considered objectives to a weighted sum (Figure
\ref{pic1}).  To overcome this problem, an approach of a
selection-operator is presented, using only few information and
providing a underlying self-adaption technique.

In this approach, we use dominance-information of the individuals of
the population by calculating for each individual $i$ the number of
alternatives $\xi_i$  from which this individual is dominated.
For a population
consisting of $n_{pop}$ alternatives we get values of:
\begin{align}
0\leq \xi_i\leq n_{pop} -1
\end{align}
Individuals that are not being dominated by others should receive a
higher fitness value than individuals that are being dominated, i.e.:

\begin{align}
\text{if } \xi_i<\xi_j \rightarrow f(i)>f(j) \quad\forall i,j=1,\dots,n_{pop}\\
\text{if } \xi_i=\xi_j \rightarrow f(i)=f(j) \quad\forall i,j=1,\dots,n_{pop}
\end{align}

We calculate the fitness value for each individual $i$ by a linear
normalization. Individuals with the lowest values of $\xi_i (\xi_i=0)$
receive the highest corresponding value of $f(i)=f_{max}$
and the individual with the
highest value $\xi_{max}=\max[\xi_i] \quad \forall i=1,\dots,n_{pop}$
receive the lowest value of $f(i)=f_{min}$.

\begin{align}
f_{max}\gg f_{min}\geq 0\\
\intertext{As a result we obtain:}
f(i)=f_{max} - \left( \frac{f_{max}-f_{min}}{\xi_{max}} \right)*\xi_i
\end{align}

\section{The implementation \cite{Geiger1997}}

The approach of the genetic algorithm is implemented in a computer
program which solves vehicle routing problems with time windows under
multiple objectives \cite{Gaskell1967}.

\begin{itemize}
\item[] The examined objectives are:
\item Minimizing the total distances traveled by the vehicles.
\item Minimizing the number of vehicles used.
\item Minimizing the time window violation.
\item Minimizing the number of violated time windows.
\end{itemize}

\begin{figure}%[h]
\begin{center}
\includegraphics[width=.95\linewidth]{\micfilename{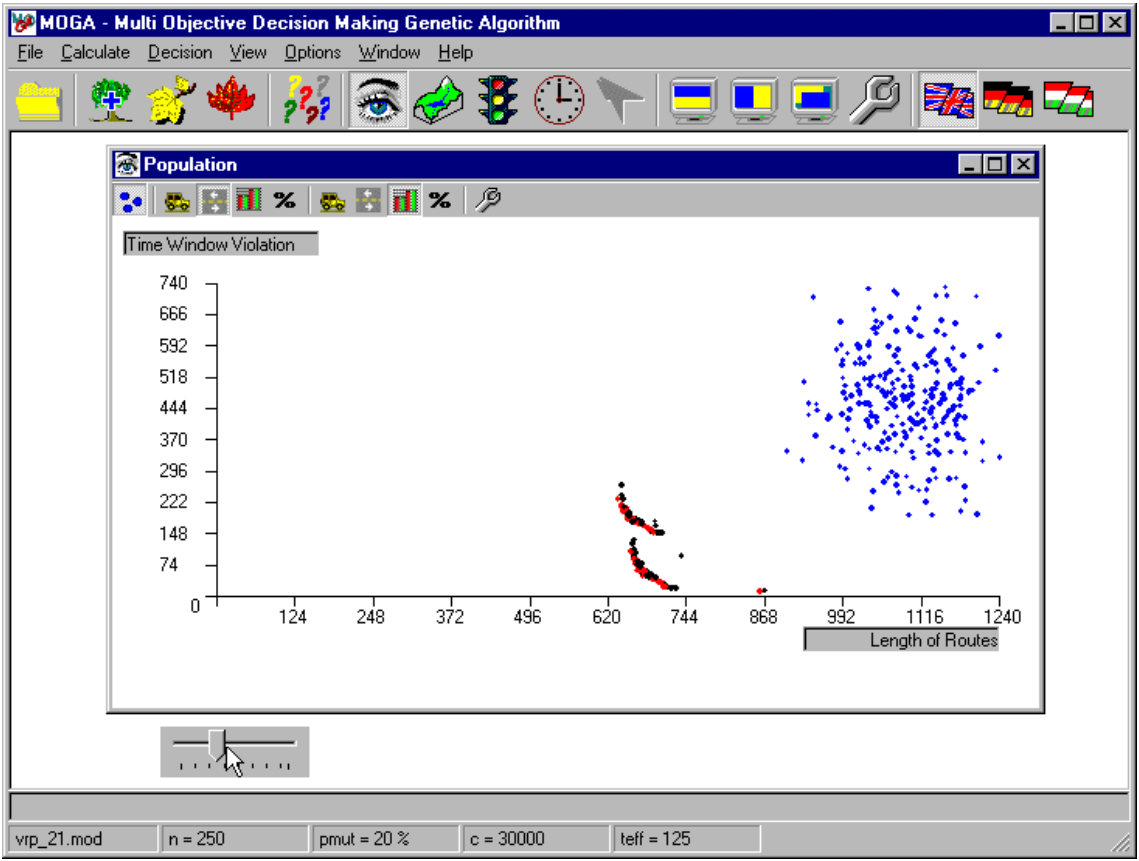}}   % Please use command
\vspace{-.5em}
\caption{Progress of the genetic algorithm}     % \micfilename
\label{pic2}
\end{center}
\end{figure}

The program illustrates the progress of the genetic algorithm and the
parameters of the approach of the can simply be controlled by the
graphical user interface (Figure \ref{pic2}).

\begin{figure}%[htb]
\begin{center}
\includegraphics[width=89mm]{\micfilename{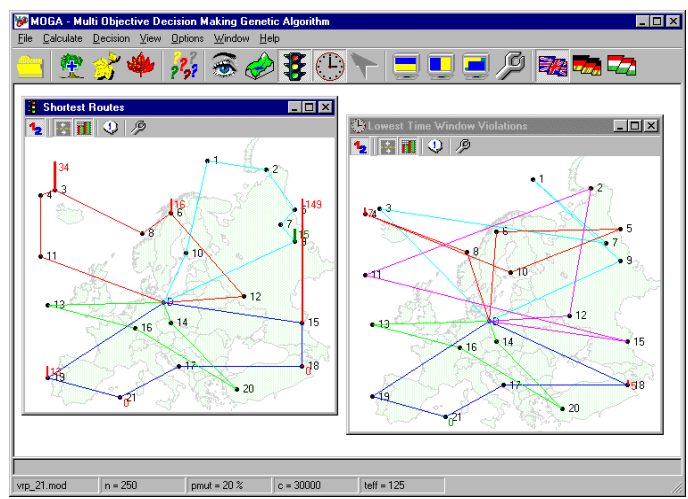}}   % Please use command
\vspace{-.5em}
\caption{Comparison of obtained alternatives}                                            % \micfilename
\label{pic3}
\end{center}
\end{figure}

In addition to the necessary calculations, the obtained alternatives
of the vehicle routing problem can easily be compared, as shown in
Figure \ref{pic3}.

For example the alternative with the shortest routes is compared to
the alternative having the lowest time window violations.  The windows
show the routes, travelled by the vehicles from the depot to the
customers. The time window violations are visualized with vertical
bars at each customer.  Red: The vehicle is too late, green: the truck
arrives too early.

\begin{figure}%[ht]
\begin{center}
\includegraphics[width=100mm]{\micfilename{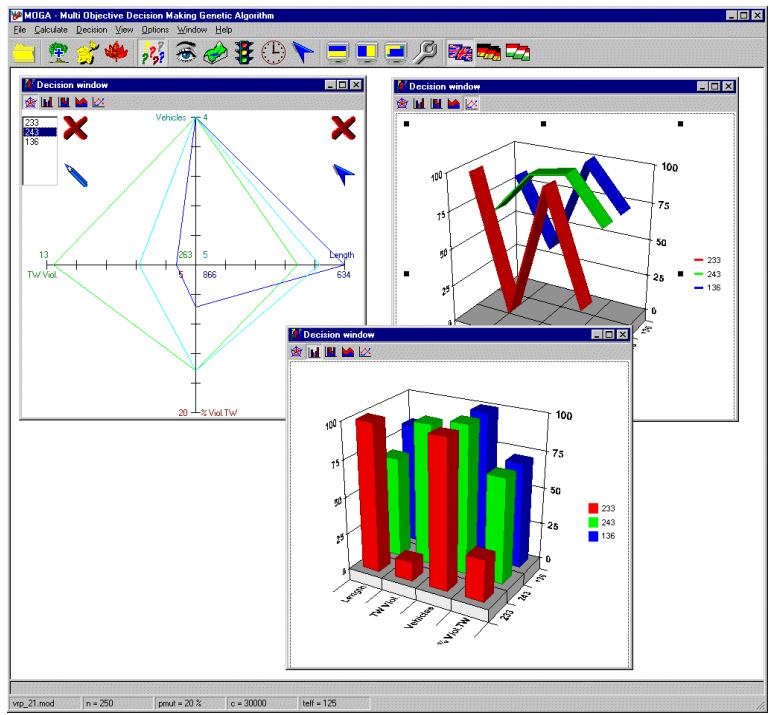}}   % Please use command
\vspace{-.5em} \caption{Decision support mode, showing trade-offs}                                            % \micfilename
\label{pic4}
\end{center}
\end{figure}

For a more detailed comparison, inverse radar charts and 3D-views are
available, showing the trade-off between the objective values of the
selected alternatives (Figure \ref{pic4}).

%\nocite{BackSchwefel1996}
%\nocite{Davis1991}
%\nocite{DesrochersLenstraSavelsbergh1990}
%\nocite{FonsecaFleming1993}
%\nocite{FonsecaFleming1995}
%\nocite{Gaskell1967}
%\nocite{Geiger1997}
%\nocite{Horn1997}
%\nocite{Schaffer1985}
%\nocite{TamakiKitaKobayashi1996}

%inclusion of bibliography trough bibtex
\bibliographystyle{plain}
%\bibliography{YourBibliographyFile}

% If not using bibtex you should comment the previous line
% (\bibliography{YourBibliographyFile} and have something like this:

\begin{thebibliography}{10}

\bibitem{BackSchwefel1996}
T.~Bäck and H.-P. Schwefel.
\newblock Evolutionary computation: An overview.
\newblock In {\em Proc. 1996 IEEE Int. Conf. Evolutionary Computation}, pages
  20--29, Piscataway NJ, 1996. IEEE Service Center.

\bibitem{Davis1991}
L.~Davis.
\newblock A genetic algorithms tutorial.
\newblock In L.~Davis, editor, {\em Handbook of genetic algorithms}, pages
  1--101. 1991.

\bibitem{DesrochersLenstraSavelsbergh1990}
M.~Desrochers, J.K. Lenstra, and M.W.P. Savelsbergh.
\newblock A classification scheme for vehicle routing and scheduling problems.
\newblock {\em European J. Oper. Res.}, 46:322--332, 1990.

\bibitem{FonsecaFleming1993}
C.M. Fonseca and P.J. Fleming.
\newblock Genetic algorithms for multiobjective optimization: formulation,
  discussion and generalization.
\newblock In Forrest S., editor, {\em Genetic Algorithms: Proc. 5-th Int.
  Conf.}, pages 416--423. Morgan Kaufmann, 1993.

\bibitem{FonsecaFleming1995}
C.M. Fonseca and P.J. Fleming.
\newblock An overview of evolutionary algorithms in multi-objective
  optimization.
\newblock {\em Evolutionary Computation}, 3:1--16, 1995.

\bibitem{Gaskell1967}
T.J. Gaskell.
\newblock Bases for vehicle fleet scheduling.
\newblock {\em Oper. Res. Quart.}, 18:281--295, 1967.

\bibitem{Geiger1997}
M.J. Geiger.
\newblock Konzeption eines genetischen algorithmus zur lösung multikriterieller
  optimierungsprobleme.
\newblock Department of Productions and Operations Management, Institute 510 -
  Business Administration, University of Hohenheim, Germany.
\newblock (in German).

\bibitem{Horn1997}
J.~Horn.
\newblock Multicriterion decision making.
\newblock In Bäck T. and Fogel D.B., editors, {\em Handbook of evolutionary
  computation}, pages 1--15. IOP Publ. \& Exford Press, New York, 1997.

\bibitem{Schaffer1985}
J.D. Schaffer.
\newblock Multiple objective optimization with vector evaluated genetic
  algorithms.
\newblock In Grefenstette J.J., editor, {\em Genetic Algorithms and Their
  Applications: Proc. 1-th Int. Conf. Genetic Algorithms}, pages 93--100,
  Hillsdale NJ, 1985.

\bibitem{TamakiKitaKobayashi1996}
H.~Tamaki, H.~Kita, and S.~Kobayashi.
\newblock Multi-objective optimization by genetic algorithms: A review.
\newblock In {\em Proc. 1996 IEEE Int. Conf. Evolutionary Computation}, pages
  517--522, Piscataway NJ, 1996. IEEE Service Center.

\end{thebibliography}
%
% \begin{thebibliography}{1}
%
% \bibitem{Dantzig63}
% George~B. Dantzig.
% \newblock {\em Linear programming and extensions}.
% \newblock Princeton University Press, Princeton, N.J., 1963.
%
% \bibitem{Lamport94}
% Leslie Lamport.
% \newblock {\em LaTeX: A Document Preparation System}.
% \newblock Addison-Wesley, 2nd edition, 1994.
%
% \bibitem{Sawyer2001}
% {ToM} Sawyer and Huckleberry Finn.
% \newblock The {TWAIN} heuristic.
% \newblock In Gutenberg, editor, {\em Proceedings of MIC'2001}, pages 111--117,
%   2001.
%
% \bibitem{Shannon48}
% C.~Shannon.
% \newblock A mathematical theory of communication.
% \newblock {\em Bell System Tech. J.}, 27:379--423, July 1948.
%
% \end{thebibliography}

%%%%%%%%%%%%%%%%%%%%%%%%%%%%%%%%%%%%%%%%%%%%%%%%%%%%%%%%%%%%%%%%%%%%%%%%%%%
%
%      Your extended abstract ends here!
%
%%%%%%%%%%%%%%%%%%%%%%%%%%%%%%%%%%%%%%%%%%%%%%%%%%%%%%%%%%%%%%%%%%%%%%%%%%%

\end{document}